\pdfoutput=1

\documentclass[11pt]{article}

\usepackage{emnlp2021}
\usepackage{booktabs}
\usepackage{times}
\usepackage{latexsym}
\usepackage{amsmath}
\usepackage{comment}
\usepackage{tikz}
\usepackage{tkz-tab}
\usepackage{caption}
\usepackage{latexsym}
\usepackage{amssymb}
\usepackage{subcaption} 
\usepackage[T1]{fontenc}

\usepackage[utf8]{inputenc}

\usepackage{microtype}

\usepackage{longtable}

\newcommand{\Ni}{({\em i})~}
\newcommand{\Nii}{({\em ii})~}
\newcommand{\Niii}{({\em iii})~}
\newcommand{\Niv}{({\em iv})~}

\usepackage{pgfplots} 
\usepackage{enumitem}

\title{Comparing Feature-Engineering and Feature-Learning Approaches \\ for Multilingual Translationese Classification}

\author{Daria Pylypenko\textsuperscript{{\normalfont *1}}, Kwabena Amponsah-Kaakyire\textsuperscript{{\normalfont *1,2}}, Koel Dutta Chowdhury\textsuperscript{{\normalfont *1}}, \\ {\bf Josef van Genabith}\textsuperscript{1,2}{\bf,} \and {\bf Cristina Espa\~{n}a-Bonet\textsuperscript{{\normalfont 2}}} \\  
  \textsuperscript{1}Saarland University,
  \textsuperscript{2}German Research Center for Artificial Intelligence (DFKI) \\
  Saarland Informatics Campus, Saarbrücken, Germany
  \\
  {\tt \{daria.pylypenko,koel.duttachowdhury\}@uni-saarland.de}\\
  {\tt kwabena.amponsah-kaakyire@dfki.de}\\ 
  {\tt \{cristinae, Josef.Van\_Genabith\}@dfki.de }\\
  } 

\begin{document}
\maketitle

\begin{abstract}
\renewcommand{\thefootnote}{\fnsymbol{footnote}}
\footnotetext[1]{Equal contribution.}Traditional hand-crafted linguistically-informed features have often been used for distinguishing between translated and original non-translated texts. By contrast, to date, neural architectures without manual feature engineering have been less explored for this task. In this work, we \Ni compare the traditional feature-engineering-based approach to the feature-learning-based one and \Nii analyse the neural architectures in order to investigate how well the hand-crafted features explain the variance in the neural models' predictions. We use pre-trained neural word embeddings, as well as several end-to-end neural architectures in both monolingual and multilingual settings and compare them to feature-engineering-based SVM classifiers. We show that \Ni neural architectures outperform other approaches by more than 20 accuracy points, with the BERT-based model performing the best in both the monolingual and multilingual settings; \Nii while many individual hand-crafted translationese features correlate with
neural model predictions, feature importance analysis shows that the most important features for neural and classical architectures differ; and  \Niii our multilingual experiments provide empirical evidence for translationese universals across languages.

\end{abstract}

\section{Introduction}
Texts originally written in a language exhibit properties that distinguish them from texts that are the result of a translation into the same language. These properties are referred to as \emph{translationese} \citep{gellerstam:1986}. Earlier studies have shown that using various hand-crafted features for supervised learning
 can be effective for translationese classification \citep{Baroni2006ANA,volanskyEtAl:2015,rubinoEtAl:2016}.  
 However, this approach has a number of limitations. Firstly, manually designed features may be partial and non-exhaustive in a sense that they are based on our linguistic intuitions, and thus may not be guaranteed  to capture all discriminative characteristics of the input data seen during training. 
Other limitations are related to the difficulties in obtaining linguistic annotation tools (e.g., parsers, taggers, etc.) for some languages, reliance on $n$-gram counts, limited contexts, corpus specific characteristics, among others. In this work, we compare a standard approach based on hand-crafted features with automatic feature learning based on data, task and learner without prior linguistic assumptions.

Moreover, most previous approaches have focused on classifying translationese in the monolingual setting,  
i.e. translations come from one or multiple source languages, but the language on which to perform the classification is always the same.
To the best of our knowledge, the multilingual setting with multiple source and target languages has not been explored yet. If translationese features are language-independent or shared among languages, multilingual translationese classification experiments would show the effect.  
 We perform binary translationese classification not only in mono-, but also in multilingual settings to empirically verify the existence of translationese universals throughout different source and target languages. 

\medskip
\noindent
In our work we investigate:

\smallskip
\Ni How automatic neural feature learning approaches to translationese classification compare to classical feature-engineering-based approaches on the same data. To do this, we use pre-trained embeddings as well as several end-to-end neural architectures.

\smallskip
\Nii Whether it is possible to effectively detect translationese in multilingual multi-source data, and how it compares to detecting translation in monolingual and single-source data in different languages.

\smallskip
\Niii Whether $a)$ translationese features learned in one setting can be useful in a different setting and $b)$ the overhead of training separate monolingual models can be reduced by either multi-source monolingual models for a given target language or even better, a multilingual model. For this we perform cross-data evaluation.

\smallskip
\Niv Whether variation observed in predictions of neural models can be explained by linguistically inspired hand-crafted features. We perform linear regression experiments to study the correlation between hand-crafted features and predictions of representation learning models as a starting point for investigating neural models which do not lend themselves easily to full explainability.

\medskip
\noindent
We show that:
\begin{itemize}
    \item representation-learning approaches outperform hand-crafted feature-selection methods for translationese classification, with BERT giving the highest accuracy,
    
    \item it is possible to classify translationese in the multilingual data, but models trained on monolingual single-source data generally yield better performance than models trained on multi-source and multilingual data,
    
    \item in contrast to hand-crafted feature-based models, neural models perform relatively well on different datasets (cross-data evaluation), and single-source can, to a reasonable extent, be substituted by multi-source mono- and multilingual models,
    
    \item many traditional hand-crafted translationese features exhibit significant correlation with the predictions of the neural models. However, a feature importance analysis  shows that the most important features for neural networks and for classical architectures differ.
\end{itemize}

The paper is organized as follows. Section~\ref{related_work} describes related work. Section~\ref{method} introduces the architectures used in our study. Section~\ref{experiments} discusses the data and presents the main classification results. We perform cross-data evaluations in Section~\ref{cross} and analyze feature importance and correlation in Section~\ref{analysis}. Finally, we summarize and draw conclusions in Section~\ref{conclusion}.

\section{Related Work}
\label{related_work}
Recent work on translationese, both on human- and machine-translated texts, explores topics ranging from translationese characterization \cite{volanskyEtAl:2015,bogoychev:2019,bizzoniEtal:2020} to unsupervised classification \cite{rabinovichWintner:2015}, to
exploring insights into the structure of language typologies with respect to different translationese properties \cite{rabinovich2017found, bjerva2019language, chowdhury2020understanding,DuttaEtal:RANLP:2021},
to the effects on downstream tasks such as machine translation \cite{stymne-2017-effect,toral-etal-2018-attaining,zhangToral:2019,freitagEtAl:2019,edunov-etal-2020-evaluation,rileyEtAl:2020, graham-etal-2020-statistical}, to translationese data collection \citep{haifa_corpus,nisioi-etal-2016-corpus,europarl-motra21}.

Traditional translationese classification approaches rely on manually designed features, such as
$n$-gram frequencies on tokens, part-of-speech (POS) tags or lemmas \citep{Baroni2006ANA,van-halteren-2008-source, Kurokawa:2009}, function word frequencies \citep{koppel-ordan-2011-translationese, tolochinsky:2018}, character-level features \citep{popescu-2011-studying, Avner2016IdentifyingTA}, surface and lexical features \citep{ilisei:2010, volanskyEtAl:2015}, syntactic features \citep{ilisei:2010, rubinoEtAl:2016}, morpheme-based features \citep{Avner2016IdentifyingTA,volanskyEtAl:2015}, information-density-based features \citep{rubinoEtAl:2016}, etc.

By contrast, to date neural approaches to translationese \citep{bjerva2019language, chowdhury2020understanding} have received less attention. While \citet{bjerva2019language} have used learned language representations to show that the distance in the representational space reflects language phylogeny, \citet{chowdhury2020understanding, DuttaEtal:RANLP:2021} use divergence from isomorphism between embedding spaces to reconstruct phylogenetic trees from translationese data. \citet{sominsky-wintner-2019-automatic} train a BiLSTM for translation direction identification and report accuracy up to 81.0\% on Europarl data.

\section{Architectures}
\label{method}

\subsection{Feature-Selection-Based Classification (\texttt{Handcr.+SVM})}
We employ the \textit{INFODENS} toolkit \cite{infodens} to extract hand-crafted features to train and evaluate a classifier. We use a support vector machine classifier (SVM) with linear kernel, and fit the hyperparameter $C$ on the validation set. For the choice of features, we replicate the setup from \citep{europarl-motra21}, using a 108-dimensional feature vector, inspired by the feature set described in \citep{rubinoEtAl:2016}. In particular, we use: 
\begin{itemize}
    \setlength\itemsep{-0.2em}
    \item[1.] \textbf{surface features}: average word length, syllable ratio, paragraph length.
    These surface features can be connected to the simplification hypothesis \cite{ilisei:2010,volanskyEtAl:2015}, as it is assumed that translations contain simpler shorter words than original texts.
    \item[2.] \textbf{lexical features}: lexical density, type-token ratio.
    These lexical features can also be linked to the simplification hypothesis, due to the assumption that original texts have richer vocabulary than translated ones and contain a higher proportion of content words \citep{laviosa:1998,Baker1993CorpusLA}.
    \item[3.] \textbf{unigram bag-of-PoS}:
    These features correspond to the source  interference (shining-through) hypothesis \citep{volanskyEtAl:2015}, as POS $n$-grams reflect grammatical structure, which might be altered in translations due to the influence of the source language grammar.
    \item[4.] \textbf{language modelling features}: log probabilities and perplexities with and without considering the end-of-sentence token, according to forward and backward $n$-gram language models ($n \in [1;5]$) built on tokens and POS tags. 
    It is hypothesized that the perplexity of translated texts may be increased because of simplification, explicitation and interference \cite{rubinoEtAl:2016}.
    \item[5.] \textbf{$n$-gram frequency distribution features}: percentages of $n$-grams in the paragraph occurring in each quartile ($n \in [1;5]$).
    This feature could be linked to the normalization hypothesis, according to which translated texts are expected to contain more collocations, i.e. high-frequency $n$-grams \citep{toury:1980, kenny2001lexis}. 
    
\end{itemize}

In our experiments, language models and $n$-gram frequency distributions are built on the training set. The $n$-gram language models are estimated with SRILM \cite{Stolcke:2002} and SpaCy\footnote{\url{https://spacy.io/}} is used for POS-tagging. Features are scaled by their maximum absolute values.
The full list of 108 features is given in the Appendix \ref{subsec:feat_list}.

\subsection{Embedding-based Classification}

\subsubsection{Average pre-trained embeddings + SVM (\texttt{Wiki+SVM)}}
We compute an average of all token vectors in the paragraph, and use this mean vector as a feature vector to train a SVM classifier with linear kernel. We work with the publicly available language specific 300-dimensional pre-trained Wiki word vector models trained on Wikipedia using \textit{fastText}%
\footnote{\url{https://fasttext.cc/docs/en/pretrained-vectors.html}}
 \citep{joulin2016fasttext}.

\subsubsection{Gaussian distributions for similarity-based classification (\texttt{Wiki+Gauss.+SVM})}
We follow \citet{das2015gaussian,nikolentzos2017multivariate} and \citet{gourrugaussian} and represent a text as a multivariate Gaussian distribution based on the distributed representations of its words. We perform similarity-based classification with SVMs where the kernel represents similarities between pairs of texts. We work with the same pre-trained Wikipedia embeddings as in \texttt{Wiki+SVM} for the words in the model and initialize the ones not contained in the model to random vectors.

Specifically, the method assumes that each word \texttt{w} is a sample drawn from a Gaussian distribution with mean vector $\mu$ and covariance matrix $\sigma^{2}$:
\begin{equation}
    \texttt{w} \sim \mathcal{N}(\mu,\,\sigma^{2})
\end{equation}
A text is then characterized by the average of its words and their covariance.
The similarity between texts is represented by the convex combination of the similarities of their mean vectors $\mu_{i}$ and $\mu_{j}$ and their covariances matrices $\sigma_{i}^{2}$ and $\sigma_{j}^{2}$: 
\begin{equation}
\small
    \textit{similarity = $\alpha(sim(\mu_{i},\mu_{j})) + (1-\alpha)(sim(\sigma_{i}^{2},\sigma_{j}^{2}))$}
    \label{kernel}
\end{equation}
where $\alpha$ $\in$ [0,1] and the similarities between the mean vectors and co-variances matrices are computed using cosine similarity and element-wise product, respectively. Finally, a SVM classifier is employed using the kernel matrices of Equation \ref{kernel} to perform the classification.

\subsection{Neural Classification}

\subsubsection{fastText classifier (\texttt{FT})}

\textit{fastText} \cite{joulin2016fasttext} is an efficient neural network model with a single hidden
layer. The \textit{fastText} model represents texts as a bag of words and bag of $n$-gram tokens. Embeddings are averaged to form the final feature vector. A linear transformation is applied before a hierarchical softmax function to calculate the class probabilities. Word vectors are trained from scratch on our data. 

\subsubsection{Pre-trained embeddings + FT (\texttt{Wiki+FT})}

In this model we work with the pre-trained word vectors from Wikipedia to initialize the fastText classifier. The data setting makes this directly comparable to \texttt{Wiki+SVM}, a non-neural classifier.

\subsubsection{Long short-term memory network (\texttt{LSTM})}
We use a single-layer uni-directional LSTM \citep{lstm-paper} with embedding and hidden layer with 128 dimensions. The embedding layer uses wordpiece subunits and is randomly-initialised. We pool (average) all hidden states, and pass the output to a binary linear classifier. We use a batch size of 32, learning rate of $1$$\cdot$$10^{-2}$, and Adam optimiser with \textit{Pytorch} defaults.

\subsubsection{Simplified transformer (\texttt{Simpl.Trf.})}

We use a single-layer encoder--decoder transformer with the same hyperparameters and wordpiece embedding layer as the LSTM. The architecture has no positional encodings. Instead, we introduce a simple cumulative sum-based contextualisation. The attention computation has been simplified to element-wise operations and there are no feedforward connections. A detailed description is provided in Appendix \ref{subsec:sformer}.

\subsubsection{Bidirectional Encoder Representations from Transformers (\texttt{BERT})}
 We use the BERT-base multilingual uncased model (12 layers, 768 hidden dimensions, 12 attention heads) \citep{devlinEtAl:2019}. Fine-tuning is done with the \textit{simpletransformers}\footnote{\url{github.com/ThilinaRajapakse/simpletransformers}} library. For this, the representation of the [CLS] token goes through a pooler, where it is linearly projected, and a $tanh$ activation is applied. Afterwards it undergoes dropout with probability 0.1 and is fed into a binary linear classifier. We use a batch size of 32, learning rate of $4 \cdot 10^{-5}$, and  the Adam optimiser with epsilon $1 \cdot 10^{-8}$. Models were fine-tuned on 4 GPUs.

\medskip
\noindent We design and compare our "lean" single-layer LSTM and simplified transformer models with BERT in order to investigate whether the amount of data and the complexity of the task necessitate complex and large networks.\footnote{The two lean architectures drastically decrease the number of core parameters. The number of parameters is 85\,M for BERT, 132\,k for the LSTM and 768 for the simplified transformer when the embedding layer and the
classifier, which are common to the 3 architectures, are not considered.}

\begin{table}
\centering
\small
\begin{tabular}{lccc}
\toprule
Corpus & Training & Dev.  & Test \\ \midrule
TRG--SRC & 30k & 6k & 6k \\
TRG--ALL & 30k & 6k & 6k \\
ALL--ALL\textsc{[3]} & 89k & 19k & 19k \\
ALL--ALL\textsc{[8]} & 67k & 14k & 14k \\
\bottomrule
\end{tabular}
\caption{\label{corpus-stats}
Number of paragraphs in each of the datasets. Average paragraph length is around 80 tokens.
}
\end{table}

\begin{table*}
\centering
\makebox[\textwidth][c]{
\small
\begin{tabular}{ lcccc cccc }
\toprule
& \texttt{Handcr.} & \texttt{Wiki} & \texttt{Wiki} &  \texttt{fastText} &\texttt{Wiki}  &\texttt{LSTM} &\texttt{Simpl.} &  \texttt{BERT} \\
&  \texttt{+SVM} & \texttt{+SVM}  &  \texttt{+Gauss.} & \texttt{(FT)}  &\texttt{+FT} & & \texttt{Trf.} &  \\
& &  & \texttt{+SVM}& & & & & \\
\midrule
DE--EN  & 71.5$\pm$0.0 &77.7$\pm$0.1 & 67.6$\pm$0.1& 88.4$\pm$0.0 &89.2$\pm$0.0 & 89.5$\pm$0.4 & 89.7$\pm$0.2 & \textbf{92.4$\pm$0.2} \\
DE--ES  & 76.2$\pm$0.0 &79.4$\pm$0.3 & 68.2$\pm$0.2 & 90.9$\pm$0.0 &91.9$\pm$0.0 & 91.9$\pm$0.2 & 91.6$\pm$0.2 & \textbf{94.4$\pm$0.1} \\
EN--DE  & 67.6$\pm$0.7 &72.5$\pm$0.2 & 64.5$\pm$0.2 & 85.1$\pm$0.0 &85.9$\pm$0.1 & 86.8$\pm$0.5 & 85.8$\pm$0.2 & \textbf{90.7$\pm$0.1} \\
EN--ES  & 70.1$\pm$0.2  &77.5$\pm$0.4 & 67.1$\pm$0.4 & 87.6$\pm$0.0 &88.7$\pm$0.0 & 89.1$\pm$0.3 & 89.3$\pm$0.4 & \textbf{91.9$\pm$0.4} \\
ES--DE  & 71.0$\pm$0.0 &75.7$\pm$0.4  & 70.1$\pm$0.4 & 88.4$\pm$0.0 &89.1$\pm$0.0 & 90.2$\pm$0.2 & 90.4$\pm$0.3 & \textbf{92.3$\pm$0.2} \\
ES--EN  & 66.7$\pm$0.0  &70.1$\pm$0.3 & 67.0$\pm$0.7 & 87.0$\pm$0.1 &87.9$\pm$0.0 & 88.8$\pm$0.4 & 88.4$\pm$0.2 & \textbf{91.4$\pm$0.3} \\
DE--ALL & 72.6$\pm$0.0 & 64.3$\pm$0.0 &65.1$\pm$0.1& 87.4$\pm$0.0 &  88.3$\pm$0.0 & 88.5$\pm$0.2 & 88.6$\pm$0.4 & \textbf{90.9$\pm$0.3} \\
EN--ALL & 65.3$\pm$0.0 & 64.6$\pm$0.0 & 62.5$\pm$0.1&  82.7$\pm$0.0 &84.4$\pm$0.0& 84.2$\pm$0.4 & 83.8$\pm$0.3 & \textbf{87.9$\pm$0.4} \\
ES--ALL & 67.4$\pm$0.0 & 67.3$\pm$0.0 & 66.5$\pm$0.2 &84.9$\pm$0.0 &85.9$\pm$0.0 & 87.0$\pm$0.3 & 86.9$\pm$0.3 & \textbf{89.9$\pm$0.1} \\
ALL--ALL\textsc{[3]} & 58.9$\pm$0.0  &--& -- &  85.0$\pm$0.0  & --    & 84.4$\pm$0.3 & 84.5$\pm$0.2 & \textbf{89.6$\pm$0.2} \\
ALL--ALL\textsc{[8]} & 65.4$\pm$0.1 & -- & -- & 70.4$\pm$0.1 & -- & 77.2$\pm$0.3 & 77.9$\pm$0.1 & \textbf{84.6$\pm$0.2} \\
\bottomrule
\end{tabular}}
\caption{\label{table4}
Translationese classification average accuracy on the mono- and multilingual test sets (average and standard deviation over 5 runs). 
}
\end{table*}
\section{Translationese Classification}
\label{experiments}

\subsection{Data}
\label{data}
We use monolingual and multilingual translationese corpora from \citet{europarl-motra21} which contain annotated paragraphs (avg. 80 tokens) of the proceedings of the European parliament, the Multilingual Parallel Direct Europarl%
\footnote{\url{github.com/UDS-SFB-B6-Datasets/Multilingual-Parallel-Direct-Europarl}} (MPDE).
Annotations indicate the source (SRC) and target languages (TRG), the "original" or "translationese" label, and whether the translations are direct or undefined (possibly translated through a pivot language). As texts translated through a pivot language may have different characteristics from directly translated texts, here we only use the direct translations. For the initial experiment we focus on 3 languages: German (DE), English (EN) and Spanish (ES). We adopt the following format for data description: we refer to translationese corpora (i.e. corpora where half of the data is originals, half translationese) with the "TRG--SRC" notation (with a dash): TRG is the language of the corpus, SRC is the source language, from which the translation into the TRG language was done in order to produce the translationese half. The "TRG$\leftarrow$SRC" notation (with an arrow) denotes the result of translating a text from SRC into TRG language. We use it to refer only to the translationese half of the corpus.

For our experiments we extract four datasets from MPDE with summary statistics in Table \ref{corpus-stats}.

\begin{enumerate}
    \item Monolingual single-source data: DE--EN, DE--ES, EN--DE, EN--ES, ES--DE, ES--EN.  For each corpus, there is an equal number of translated and original paragraphs. 

    \item Monolingual multi-source data: DE--ALL, EN--ALL, ES--ALL. For DE--ALL, e.g., half of the data is DE original texts, and the other half contains equal proportions of DE$\leftarrow$ES and DE$\leftarrow$EN.

    \item Multilingual multi-source data: ALL--ALL\textsc{[3]}. There is an equal number of originals: DE, EN and ES, which together make up 50\% of the examples. The other 50\% which are translated are equal proportions of DE$\leftarrow$EN, DE$\leftarrow$ES, EN$\leftarrow$DE, EN$\leftarrow$ES, ES$\leftarrow$DE and ES$\leftarrow$EN. 
\end{enumerate}

EN, DE and ES are relatively close typologically. We conduct additional experiments in order to investigate how well the classification can be performed when more and more distant languages are involved: 

\begin{enumerate}[resume]
    \item Multilingual multi-source data large: ALL--ALL\textsc{[8]}, 
    balanced in the same way as ALL--ALL\textsc{[3]}, but with the addition of Greek (EL), French (FR), Italian (IT), Dutch (NL) and Portuguese (PT).
\end{enumerate}

\begin{table*}
\centering
\small
\begin{tabular}{lccccccccc}
\toprule
& DE--EN & DE--ES & EN--DE & EN--ES & ES--DE & ES--EN & DE--ALL & EN--ALL & ES--ALL \\
\midrule
DE--EN & 92.4$\pm$0.2 & 76.6$\pm$0.7 & - & - & - & - & 90.5$\pm$0.3 & - & - \\
DE--ES & 82.6$\pm$1.1 & 94.4$\pm$0.1 & - & - & - & - & 91.8$\pm$0.4 & - & - \\
EN--DE & - & - & 90.7$\pm$0.1 & 64.7$\pm$1.4 & - & - & - & 87.3$\pm$0.4  & - \\
EN--ES & - & - & 72.9$\pm$0.9 & 91.9$\pm$0.4 & - & - & - & 88.6$\pm$0.4 & - \\
ES--DE & - & - & - & - & 92.3$\pm$0.2 & 78.8$\pm$0.9 & - & - & 90.6$\pm$0.1 \\
ES--EN & - & - & - & - & 78.8$\pm$1.6 & 91.4$\pm$0.3 & - & - & 89.0$\pm$0.2 \\
DE--ALL & 87.3$\pm$0.6 & 85.3$\pm$0.4 & - & - & - & - & 90.9$\pm$0.3 & - & - \\
EN--ALL & - & - & 81.7$\pm$0.5 & 78.3$\pm$0.7 & - & - & - & 87.9$\pm$0.4 & - \\
ES--ALL & - & - & - & - & 85.9$\pm$0.9 & 85.0$\pm$0.6 & - & - & 89.9$\pm$0.1 \\
\bottomrule
\end{tabular}
\caption{\label{cross-mono-bert}
BERT translationese classification accuracy of all TRG--SRC and TRG--ALL models on TRG--SRC and TRG--ALL test sets (average and standard deviation over 5 runs). Columns: training set; rows: test set. 
}
\end{table*}

For all settings we perform binary classification: original vs. translated.

\subsection{Results}
\label{result}

 Paragraph-level translationese classification results with mean and standard deviations over 5 runs are reported in Table \ref{table4}. Overall, the \texttt{BERT} model outperforms other architectures in all settings, followed closely by the other end-to-end neural architectures. Using the pre-trained Wiki embeddings helps improving the accuracy of the \texttt{fastText} method in all cases. Among the approaches with the SVM classifier, \texttt{Wiki+SVM} performs best in the single-source settings, but shows lower accuracy than \texttt{Handcr.+SVM} in the multi-source (TRG–ALL) settings.
\texttt{Wiki+Gauss.+SVM} performs worst apart from on ES--EN and DE--ALL.

In the monolingual single-source settings, we observe that accuracy is slightly lower when the source language is typologically closer to the text language, i.e. it becomes more difficult to detect translationese. Specifically, DE--EN tends to have lower accuracy than DE--ES; EN--DE lower accuracy than EN--ES; and ES--EN lower accuracy than ES--DE. Accuracy generally drops when going from single-source to the multi-source setting, e.g. from DE--EN and DE--ES to DE--ALL. The EN--ALL dataset is the most difficult for most of the models among the TRG--ALL datasets. The ALL--ALL\textsc{[3]} setting exhibits comparable accuracy to the TRG--ALL setting for the neural models, but for the SVM there is a drop of around 9 points. Throughout our discussion we always report absolute differences between systems. The ALL--ALL\textsc{[8]} data results in reduced accuracy for most architectures, except \texttt{Handcr.+SVM}.

Neural-classifier-based models substantially outperform the other architectures: the SVMs trained with hand-crafted linguistically-inspired features, e.g., trail BERT by $\sim$20 accuracy points.

To make sure our hand-crafted-feature-based SVM results are competitive, we compare them with \citet{rabinovichWintner:2015} on our data. \citet{rabinovichWintner:2015} show that training a SVM classifier on the top 1000 most frequent POS- or character-trigrams yields SOTA translationese classification results on Europarl data. On our data, POS-trigrams yield around 5 points increase in accuracy for most of the datasets and character-trigrams tend to lower the accuracy by around 4 points (Appendix~\ref{subsec:baselines}). For the remainder of the paper we continue to work with our hand-crafted features, designed to capture various linguistic aspects of translationese.

\begin{table*}[!h]
\centering
\small
\begin{tabular}{lccccc}
\toprule
& \texttt{Handcr.} &\texttt{fastText}  & \texttt{Simpl.} & \texttt{LSTM} & \texttt{BERT} \\
& \texttt{+SVM} &  \texttt{(FT) }   &  \texttt{Trf.} & \\
\midrule

DE--EN & 58.5$\pm$0.0 ($\downarrow$13.0) & 85.9$\pm$0.0 ($\downarrow$2.5) & 85.5$\pm$0.5 ($\downarrow$4.3) & 86.6$\pm$0.7 ($\downarrow$2.9) & 90.5$\pm$0.3 ($\downarrow$1.9) \\
DE--ES & 57.0$\pm$0.0 ($\downarrow$19.2) 
&88.3$\pm$0.0 ($\downarrow$2.6) & 87.2$\pm$0.5 ($\downarrow$4.3) & 85.3$\pm$0.3 ($\downarrow$6.9) & 91.5$\pm$0.2 ($\downarrow$2.9) \\
EN--DE & 50.0$\pm$0.0 ($\downarrow$17.6)
&81.5$\pm$0.1 ($\downarrow$3.6) & 81.1$\pm$0.3 ($\downarrow$4.7) & 80.9$\pm$0.3 ($\downarrow$5.8) & 87.2$\pm$0.4 ($\downarrow$3.5) \\
EN--ES & 50.5$\pm$0.0 ($\downarrow$19.6) & 84.6$\pm$0.0 ($\downarrow$3.0) & 83.5$\pm$0.5 ($\downarrow$5.9) & 83.8$\pm$0.6 ($\downarrow$5.3) & 88.9$\pm$0.3 ($\downarrow$3.0) \\
ES--DE & 50.0$\pm$0.0 ($\downarrow$21.0) &  85.6$\pm$0.0 ($\downarrow$2.8) & 86.2$\pm$0.4 ($\downarrow$4.3) & 85.7$\pm$0.5 ($\downarrow$4.6) & 90.4$\pm$0.4 ($\downarrow$1.9) \\
ES--EN & 51.3$\pm$0.0 ($\downarrow$15.4) & 84.1$\pm$0.0 ($\downarrow$2.9) & 84.6$\pm$0.4 ($\downarrow$0.4) & 82.1$\pm$0.4 ($\downarrow$6.7) & 89.0$\pm$0.4 ($\downarrow$2.4) \\
DE--ALL & 59.9$\pm$0.0 ($\downarrow$12.7) & 87.2$\pm$0.0 ($\downarrow$0.2) & 86.3$\pm$0.4 ($\downarrow$2.3) & 85.9$\pm$0.5 ($\downarrow$2.6) & 90.8$\pm$0.1 ($\downarrow$0.1) \\
EN--ALL & 50.2$\pm$0.0 ($\downarrow$15.1) &82.9$\pm$0.0 ($\uparrow$0.2) & 82.0$\pm$0.1 ($\downarrow$1.8) & 82.2$\pm$0.2 ($\downarrow$2.1) & 88.1$\pm$0.5 ($\uparrow$0.2) \\
ES--ALL & 50.0$\pm$0.0 ($\downarrow$17.4) & 84.8$\pm$0.0 ($\downarrow$0.1) & 85.3$\pm$0.2 ($\downarrow$1.6) & 85.2$\pm$0.5 ($\downarrow$1.8) & 89.8$\pm$0.3 ($\downarrow$0.1) \\
ALL--ALL\textsc{[3]} & 58.9$\pm$0.0 (0.0) & 85.0$\pm$0.0 (0.0) & 84.5$\pm$0.2 (0.0) & 84.4$\pm$0.3 (0.0) & 89.6$\pm$0.2 (0.0) \\
\bottomrule
\end{tabular}
\caption{\label{cross-multi}
Translationese classification accuracy of the ALL--ALL\textsc{[3]} model on all test sets (average and standard deviations over 5 runs). The difference from actual trained model performance is indicated in parentheses. 
}
\end{table*}

\begin{table*}
\centering
\small
\begin{tabular}{lccccc}
\toprule
& \texttt{Handcr.} &\texttt{fastText}  & \texttt{Simpl.} & \texttt{LSTM} & \texttt{BERT} \\
& \texttt{+SVM} &  \texttt{(FT) }   &  \texttt{Trf.} & \\
\midrule

DE--EN & 53.0$\pm$0.5 ($\downarrow$18,5) & 71.0$\pm$0.3 ($\downarrow$17.4) & 79.3$\pm$0.4 ($\downarrow$12.3) & 79.9$\pm$0.5 ($\downarrow$9.6) & 85.5$\pm$0.4 ($\downarrow$6.9) \\
DE--ES & 51.3$\pm$0.3 ($\downarrow$24.9) & 73.2$\pm$0.3 ($\downarrow$17.7) & 81.4$\pm$0.3 ($\downarrow$8.4) & 79.0$\pm$0.5 ($\downarrow$12.9) & 87.9$\pm$0.3 ($\downarrow$6.5) \\
EN--DE & 48.3$\pm$0.1 ($\downarrow$19.3) & 65.8$\pm$0.2 ($\downarrow$19.3) & 74.2$\pm$1.0 ($\downarrow$11.6) & 72.9$\pm$0.4 ($\downarrow$13.8) & 79.0$\pm$0.5 ($\downarrow$11.7) \\
EN--ES & 50.3$\pm$0.1 ($\downarrow$19.8) &68.9$\pm$0.3 ($\downarrow$18.7) & 76.8$\pm$0.6 ($\downarrow$12.8) & 75.6$\pm$0.8 ($\downarrow$13.5) & 83.2$\pm$0.4 ($\downarrow$8.7) \\
ES--DE & 50.0$\pm$0.0 ($\downarrow$21.0) & 71.1$\pm$0.2 ($\downarrow$17.3) & 78.8$\pm$0.5 ($\downarrow$11.6) & 76.0$\pm$0.7 ($\downarrow$14.2) & 83.8$\pm$0.3 ($\downarrow$8.5) \\
ES--EN & 53.2$\pm$0.5 ($\downarrow$13.5) & 69.9$\pm$0.2 ($\downarrow$17.1) & 76.7$\pm$0.6 ($\downarrow$11.7) & 75.4$\pm$0.7 ($\downarrow$13.4) & 82.8$\pm$0.2 ($\downarrow$8.6) \\
DE--ALL & 53.1$\pm$0.5 ($\downarrow$19.5) &72.1$\pm$0.3 ($\downarrow$15.3) & 80.5$\pm$0.4 ($\downarrow$8.1) & 79.7$\pm$0.6 ($\downarrow$8.8) & 86.8$\pm$0.2 ($\downarrow$4.1) \\
EN--ALL & 48.4$\pm$0.2 ($\downarrow$16.9) &67.0$\pm$0.2 ($\downarrow$15.7) & 75.4$\pm$0.9 ($\downarrow$8.4) & 74.4$\pm$0.5 ($\downarrow$9.9) & 81.1$\pm$0.1 ($\downarrow$6.8)\\
ES--ALL & 50.8$\pm$0.3 ($\downarrow$16.6) &70.4$\pm$0.2 ($\downarrow$14.5) & 77.9$\pm$0.6 ($\downarrow$9.1) & 75.9$\pm$0.6 ($\downarrow$11.1) & 83.2$\pm$0.3 ($\downarrow$6.7)\\
ALL--ALL\textsc{[3]} & 53.2$\pm$0.3 ($\downarrow$5.7) &70.5$\pm$0.2 ($\downarrow$14.5)& 77.9$\pm$0.2 ($\downarrow$6.6) & 76.7$\pm$0.5 ($\downarrow$7.7) & 83.7$\pm$0.1 ($\downarrow$5.9)\\
ALL--ALL\textsc{[8]} & 65.4$\pm$0.1 (0.0) &70.4$\pm$0.1 (0.0) & 77.9$\pm$0.1 (0.0) & 77.2$\pm$0.3 (0.0) & 84.6$\pm$0.2 (0.0) \\
\bottomrule
\end{tabular}
\caption{\label{cross-multi8}
As Table~\ref{cross-multi} for the ALL--ALL\textsc{[8]} model.
}
\end{table*}

\section{Multilinguality and Cross-Language Performance}
\label{cross}

Since neural architectures perform better than the non-neural ones, we perform the multilingual and cross-language analysis only with the neural models. We evaluate the models trained on one dataset on the other ones, in order to verify:

\begin{itemize}
    \item Whether for a given target language, the model trained to detect translationese from one source language, can detect translationese from another source language: TRG--SRC$_1$ on TRG--SRC$_2$, and TRG--SRC on TRG--ALL;
    \item How well the model trained to detect translationese from multiple source languages can detect translationese from a single source language: TRG--ALL on TRG--SRC, and ALL--ALL\textsc{[3]} on TRG--SRC;
    \item How well the model trained to detect translationese in multilingual data performs on monolingual data: ALL--ALL\textsc{[3]} on TRG--ALL, and ALL--ALL\textsc{[3]} on TRG--SRC.
\end{itemize}

Table \ref{cross-mono-bert} shows the results of cross-data testing for the monolingual models for the best-performing architecture (\texttt{BERT}). For the single-source monolingual models, we observe a relatively smaller drop (up to 13 percentage points) in performance when testing TRG--SRC on TRG--ALL (as compared to testing TRG--SRC on TRG--SRC), and a larger drop (up to 27 points) when testing TRG--SRC$_1$ on TRG--SRC$_2$ (as compared to testing TRG--SRC$_1$ on TRG--SRC$_2$).
The fact that classification performance stays above 64\% confirms the hypothesis that translationese features are source-language-independent. 

Another trend that can be observed is that in cross testing TRG--SRC$_1$ and TRG--SRC$_2$, the model where the source language is more distant from the target suffers larger performance drop when tested on the test set with the closer-related source language, than the other way around. For instance, the DE--ES model tested on the DE--EN data suffers a decrease of 17.8 points, and DE--EN model tested on the DE--ES data suffers a decrease of 9.8 points. This may be due to DE--EN having learned more of the general translationese features, which helps the model to obtain higher accuracy on the data with a different source, while the DE--ES model may have learned to rely more on the language-pair-specific features, and therefore it gives lower accuracy on the data with the different source. A similar observation has been made by \citet{koppel-ordan-2011-translationese}.

For the multi-source monolingual models (TRG--ALL), testing on TRG--SRC$_1$ and TRG--SRC$_2$ datasets shows a slight increase in performance for a source language that is more distant from the target, and a slight decrease for the more closely-related source language (as compared to testing TRG--ALL on TRG--ALL). 

Table \ref{cross-multi} displays the results of testing the multilingual (ALL--ALL\textsc[3]) models on all test sets for the neural architectures, as well as \texttt{Handcr.+SVM}. We observe that the largest performance drop (as compared to testing on ALL--ALL\textsc[3] test set) happens for the EN--DE test set. For the DE--ES set, the performance actually increases for the neural models, but not for the \texttt{Handcr.+SVM}. We extended this experiment in Table \ref{cross-multi8}, testing the ALL--ALL\textsc[8] on all test sets to further complement our multilingual analysis with more diverse languages and observe a similar trend, which is in line with the accuracy of the ALL--ALL\textsc[3] models on all test sets.

We also compare the performance of ALL--ALL[3] on different test sets to the original performance of the models trained on these datasets (in parentheses). There is a relatively larger drop in accuracy for the TRG--SRC data, than for TRG--ALL data. The largest drop for neural models is 6.7 accuracy points whilst the smallest performance drop for the \texttt{Handcr.+SVM} is 12.7. This highlights the ability of the neural models to learn features in a multilingual setting which generalize well to their component languages whereas the \texttt{Handcr.+SVM} method does not seem to work well for such a case. However, for ALL--ALL\textsc{[8]} models, Table \ref{cross-multi8} shows a large performance drop across all architectures as compared to the results from the models specifically trained for the task. The actual models are trained on language-specific features, whereas the ALL--ALL\textsc{[8]} model is trained on more diverse data containing  typologically distant languages and thus captures less targeted translationese signals. 
\medskip
\noindent 

In summary, we observe that:
\begin{itemize}
    \item For a given target language, even though a neural model trained on one source language can decently identify translationese from another source language, the decrease in performance is substantial. 
    \item Neural models trained on multiple sources for a given target language perform reasonably well on single-source languages.
    \item Neural models trained on multilingual data ALL--ALL[3] perform reasonably well on monolingual data, especially for multi-source monolingual data.
    \item Using more source and target languages (ALL--ALL\textsc{[8]}) leads to a larger decrease in cross-testing accuracy.
\end{itemize}

\section{Feature Importance and Relation to Neural Models}
\label{analysis}

\begin{figure}[t]
\centering
\begin{tikzpicture}
\pgfplotsset{%
    width=0.38\textwidth,
    every axis/.append style={
                    tick label style={font=\tiny}                      }
}
\begin{axis}[ 
    xbar, 
    xmin=0,
    symbolic y coords={{LM$_{tok}$ fwd n=3 Log Prob }, {LM$_{POS}$ bck n=1 Log Prob}, {LM$_{POS}$ fwd n=1 Log Prob}, {POS Tag Ratio Adv}, {LM$_{POS}$ bck n=2 Log Prob}, {LM$_{POS}$ fwd n=2 Log Prob}, {LM$_{tok}$ bck n=1 Log Prob}, {LM$_{tok}$ fwd n=1 Log Prob}, {POS Tag Ratio X }, {Paragraph length}},
    ytick=data,
    nodes near coords, 
    nodes near coords align={horizontal},
     enlarge y limits=0.1,
             every node near coord/.append style={/pgf/number format/.cd, fixed,
                fixed zerofill, precision=1, /tikz/.cd, font=\tiny},
     axis y line*=none,
      axis x line*=bottom,
]
\addplot[fill=blue!90,draw=black!70,tickwidth = 0pt, bar width=4pt] coordinates {(1.64,{LM$_{tok}$ fwd n=3 Log Prob }) (1.67,{LM$_{POS}$ bck n=1 Log Prob}) (1.67,{LM$_{POS}$ fwd n=1 Log Prob}) (1.83,{POS Tag Ratio Adv}) (1.91,{LM$_{POS}$ bck n=2 Log Prob}) (1.91,{LM$_{POS}$ fwd n=2 Log Prob}) (1.98,{LM$_{tok}$ bck n=1 Log Prob}) (1.98,{LM$_{tok}$ fwd n=1 Log Prob}) (2.84,{POS Tag Ratio X }) (5.07,{Paragraph length})};
    \end{axis}
\end{tikzpicture}
\caption{Top 10 SVM features, as a function of the absolute value of its feature weight.}
\label{fig:plot_svm}
\end{figure}
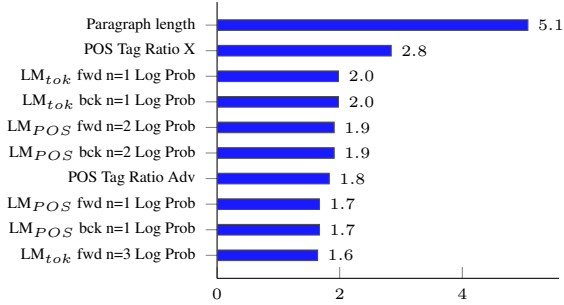

In this section we aim to quantify the feature importance of the hand-crafted linguistically inspired features used in \texttt{Handcr.+SVM} according to different multilingual models (ALL--ALL[3] setting).

As we use a Support Vector Machine with a linear kernel, we can interpret the magnitude of the feature weights as a feature importance measure. \citet{guyon} for instance, use the squared SVM weights for feature selection. We rank the features by the absolute value of the weight. The feature ranks are listed in Appendix~\ref{subsec:feat_list}. Figure \ref{fig:plot_svm} shows the top 10 features. Paragraph length is the most relevant feature, and we observe that most of the top features correspond to paragraph log probability. These features characterize simplification in translationese.

To explore whether there is any correlation between the hand-crafted features and predictions of the trained neural models, we conduct the following experiment in the multilingual setting. We fit a linear regression model for each hand-crafted feature, using the estimated probabilities of neural model as gold labels to be predicted. More formally, with $n$ paragraphs (p$_{i}$, $i=1...n$) in the test set and $d$ features, for each feature vector \textbf{x}$_j \in R^n$, $j=1...d$ we fit a model 
\begin{equation}
    \textbf{y} = w_j \textbf{x}_j  + b_j,
\end{equation}
where $w_j, b_j \in R$ are the model parameters, and \textbf{y} $\in R^n$ is a vector of predictions of the neural model $F$ (LSTM, Simplified Transformer, BERT) on the test set, with each dimension $y_i$ showing the probability of a data point to belong to the translationese class:
\begin{equation}
    \textbf{y}_i = \text{P}(F(\text{p}_i)=1)
\end{equation}

We apply min--max normalization to the features. 
We find that a large proportion of the linguistically motivated features are statistically significant for predicting the neural models' predicted probabilities, namely 60 features (out of 108) are significant for LSTM, 38 for the Simplified Transformer, and 56 for BERT, each with probability $99.9\%$. We also fit the per-feature linear models to predict the actual gold labels (and not the predictions of the neural models) to investigate which features correlate with the ground truth classes, and find 55 features to be statistically significant with $99.9\%$ probability. The full list of statistically significant features for each model, as well as for the gold labels is given in the Appendix~\ref{subsec:feat_list}. We observe that the features significant for the neural models largely overlap with the features significant for the gold labels: the $F_1$-score (as a measure of overlap) is 0.89 for LSTM, 0.75 for Simplified Transformer and 0.99 for BERT. This is expected, because high-performing neural models output probabilities that are generally close to the gold labels, therefore a similar correlation with hand-crafted features occurs.

\begin{figure*}
  \begin{minipage}{.45\textwidth}
        \begin{tikzpicture}
        \pgfplotsset{%
            width=0.7\textwidth,
            height=0.7\textwidth,
            every axis/.append style={
                            tick label style={font=\tiny}
                                           }
        }
        \begin{axis}[ 
            xbar, 
            xmin=0,
            symbolic y coords={{LM$_{POS}$ fwd n=3 Ppl}, {LM$_{POS}$ bck n=2 Ppl$_{-EOStok}$}, {LM$_{POS}$ fwd n=2 Ppl$_{-EOStok}$}, {LM$_{POS}$ fwd n=1 Ppl}, {LM$_{POS}$ bck n=1 Ppl}, {LM$_{POS}$ bck n=2 Ppl}, {LM$_{POS}$ fwd n=2 Ppl}, {POS Tag Ratio Adv}, {POS Tag Ratio Det}, {POS Tag Ratio Adp}},
            ytick=data,
            nodes near coords, 
            nodes near coords align={horizontal},
             enlarge y limits=0.1,
             every node near coord/.append style={/pgf/number format/.cd, fixed,
                fixed zerofill, precision=1, /tikz/.cd, font=\tiny},
             axis y line*=none,
              axis x line*=bottom
        ]

        \addplot[fill=blue!90,draw=black!70,tickwidth = 0pt, tick label style={font=\tiny}, bar width=4pt, label style= {format/.cd}] coordinates {(6.32,{LM$_{POS}$ fwd n=3 Ppl}) (6.96,{LM$_{POS}$ bck n=2 Ppl$_{-EOStok}$}) (6.97,{LM$_{POS}$ fwd n=2 Ppl$_{-EOStok}$}) (7.44,{LM$_{POS}$ fwd n=1 Ppl}) (7.44,{LM$_{POS}$ bck n=1 Ppl}) (8.13,{LM$_{POS}$ bck n=2 Ppl}) (8.13,{LM$_{POS}$ fwd n=2 Ppl}) (8.86,{POS Tag Ratio Adv}) (9.60,{POS Tag Ratio Det}) (9.60,{POS Tag Ratio Adp})};
        \end{axis}
        \end{tikzpicture}
    \begin{center}
    \small
    (a) BERT    
    \bigskip
    \end{center}

  \end{minipage} \quad
  \begin{minipage}{.45\textwidth}
        \begin{tikzpicture}
        \pgfplotsset{%
            width=0.7\textwidth,
            height=0.7\textwidth,
            every axis/.append style={
                            tick label style={font=\tiny}                      }
        }
        \begin{axis}[ 
            xbar, 
            xmin=0,
            symbolic y coords={{LM$_{POS}$ fwd n=1 Ppl$_{-EOS}$}, {LM$_{POS}$ bck n=2 Ppl$_{-EOS}$}, {LM$_{POS}$ fwd n=2 Ppl$_{-EOS}$}, {LM$_{POS}$ bck n=1 Ppl}, {LM$_{POS}$ fwd n=1 Ppl}, {POS Tag Ratio Adv}, {LM$_{POS}$ bck n=2 Ppl}, {LM$_{POS}$ fwd n=2 Ppl}, {POS Tag Ratio Adp}, {POS Tag Ratio Det}},
            ytick=data,
            nodes near coords, 
            nodes near coords align={horizontal},
             enlarge y limits=0.1,
             every node near coord/.append style={/pgf/number format/.cd, fixed,
                fixed zerofill, precision=1, /tikz/.cd, font=\tiny},
             axis y line*=none,
              axis x line*=bottom
        ]
        \addplot[fill=blue!90,draw=black!70,tickwidth = 0pt, tick label style={font=\tiny}, bar width=4pt] coordinates {(5.71,{LM$_{POS}$ fwd n=1 Ppl$_{-EOS}$}) (6.32,{LM$_{POS}$ bck n=2 Ppl$_{-EOS}$}) (6.32,{LM$_{POS}$ fwd n=2 Ppl$_{-EOS}$}) (6.84,{LM$_{POS}$ bck n=1 Ppl}) (6.84,{LM$_{POS}$ fwd n=1 Ppl}) (7.43,{POS Tag Ratio Adv}) (7.47,{LM$_{POS}$ bck n=2 Ppl}) (7.47,{LM$_{POS}$ fwd n=2 Ppl}) (7.60,{POS Tag Ratio Adp}) (8.48,{POS Tag Ratio Det})};
        \end{axis}
        \end{tikzpicture}
    \begin{center}
    \small
    (b) Gold labels    
    \bigskip
    \end{center}
  \end{minipage}
  \bigskip
  \hspace*{0.52em}
  \begin{minipage}{.47\textwidth}
        \begin{tikzpicture}
        \pgfplotsset{%
            width=0.7\textwidth,
            height=0.67\textwidth,
            every axis/.append style={
                            label style={font=\tiny},
                            tick label style={font=\tiny}                      }
        }
        \begin{axis}[ 
            xbar, 
            xmin=0,
            symbolic y coords={{LM$_{POS}$ bck n=1 Ppl$_{-EOS}$}, {LM$_{POS}$ fwd n=1 Ppl$_{-EOS}$}, {LM$_{POS}$ bck n=2 Ppl$_{-EOS}$}, {LM$_{POS}$ fwd n=2 Ppl$_{-EOS}$}, {POS Tag Ratio Det}, {LM$_{POS}$ bck n=2 Ppl}, {LM$_{POS}$ fwd n=2 Ppl}, {LM$_{POS}$ bck n=1 Ppl}, {LM$_{POS}$ fwd n=1 Ppl}, {POS Tag Ratio Adv}},
            ytick=data,
            nodes near coords, 
            nodes near coords align={horizontal},
             enlarge y limits=0.1,
             every node near coord/.append style={/pgf/number format/.cd, fixed,
                fixed zerofill, precision=1, /tikz/.cd, font=\tiny},
             axis y line*=none,
              axis x line*=bottom
        ]
        \addplot[fill=blue!90,draw=black!70,tickwidth = 0pt, tick label style={font=\tiny}, bar width=4pt] coordinates {(7.292,{LM$_{POS}$ bck n=1 Ppl$_{-EOS}$}) (7.292,{LM$_{POS}$ fwd n=1 Ppl$_{-EOS}$}) (7.647,{LM$_{POS}$ bck n=2 Ppl$_{-EOS}$}) (7.648,{LM$_{POS}$ fwd n=2 Ppl$_{-EOS}$}) (7.989,{POS Tag Ratio Det}) (9.745,{LM$_{POS}$ bck n=2 Ppl}) (9.746,{LM$_{POS}$ fwd n=2 Ppl}) (10.269,{LM$_{POS}$ bck n=1 Ppl}) (10.269,{LM$_{POS}$ fwd n=1 Ppl}) (14.044,{POS Tag Ratio Adv})};
        \end{axis}
        \end{tikzpicture}
    \begin{center}
    \small
    (c) Simplified Transformer    
    \end{center}

  \end{minipage} 
  \hspace{0.8cm}
  \begin{minipage}{.45\textwidth}
        \begin{tikzpicture}
        \pgfplotsset{%
            width=0.7\textwidth,
            height=0.7\textwidth,
            every axis/.append style={
                            label style={font=\tiny},
                            tick label style={font=\tiny}                      }
        }
        \begin{axis}[ 
            xbar, 
            xmin=0,
            symbolic y coords={{LM$_{POS}$ bck n=1 Ppl$_{-EOS}$}, {LM$_{POS}$ fwd n=1 Ppl$_{-EOS}$}, {LM$_{POS}$ bck n=2 Ppl$_{-EOS}$}, {LM$_{POS}$ fwd n=2 Ppl$_{-EOS}$}, {LM$_{POS}$ bck n=1 Ppl}, {LM$_{POS}$ fwd n=1 Ppl}, {LM$_{POS}$ bck n=2 Ppl}, {LM$_{POS}$ fwd n=2 Ppl}, {POS Tag Ratio Adv}, {POS Tag Ratio Det}},
            ytick=data,
            nodes near coords, 
            nodes near coords align={horizontal},
             enlarge y limits=0.1,
             every node near coord/.append style={/pgf/number format/.cd, fixed,
                fixed zerofill, precision=1, /tikz/.cd, font=\tiny},
             axis y line*=none,
              axis x line*=bottom
        ]
        \addplot[fill=blue!90,draw=black!70,tickwidth = 0pt, tick label style={font=\tiny}, bar width=4pt] coordinates
 {(7.555,{LM$_{POS}$ bck n=1 Ppl$_{-EOS}$}) (7.555,{LM$_{POS}$ fwd n=1 Ppl$_{-EOS}$}) (8.282,{LM$_{POS}$ bck n=2 Ppl$_{-EOS}$}) (8.283,{LM$_{POS}$ fwd n=2 Ppl$_{-EOS}$}) (8.92,{LM$_{POS}$ bck n=1 Ppl}) (8.92,{LM$_{POS}$ fwd n=1 Ppl}) (9.474,{LM$_{POS}$ bck n=2 Ppl}) (9.474,{LM$_{POS}$ fwd n=2 Ppl}) (10.477,{POS Tag Ratio Adv}) (11.0020,{POS Tag Ratio Det})};
        \end{axis}
        \end{tikzpicture}
    \begin{center}
    \small
    (d) LSTM    
    \end{center}

  \end{minipage}
\caption{Top 10 features as a function of $R^2 \cdot 10^{-3}$ for the neural architectures and the gold labels.}
\label{fig:plot_custom}

\end{figure*}
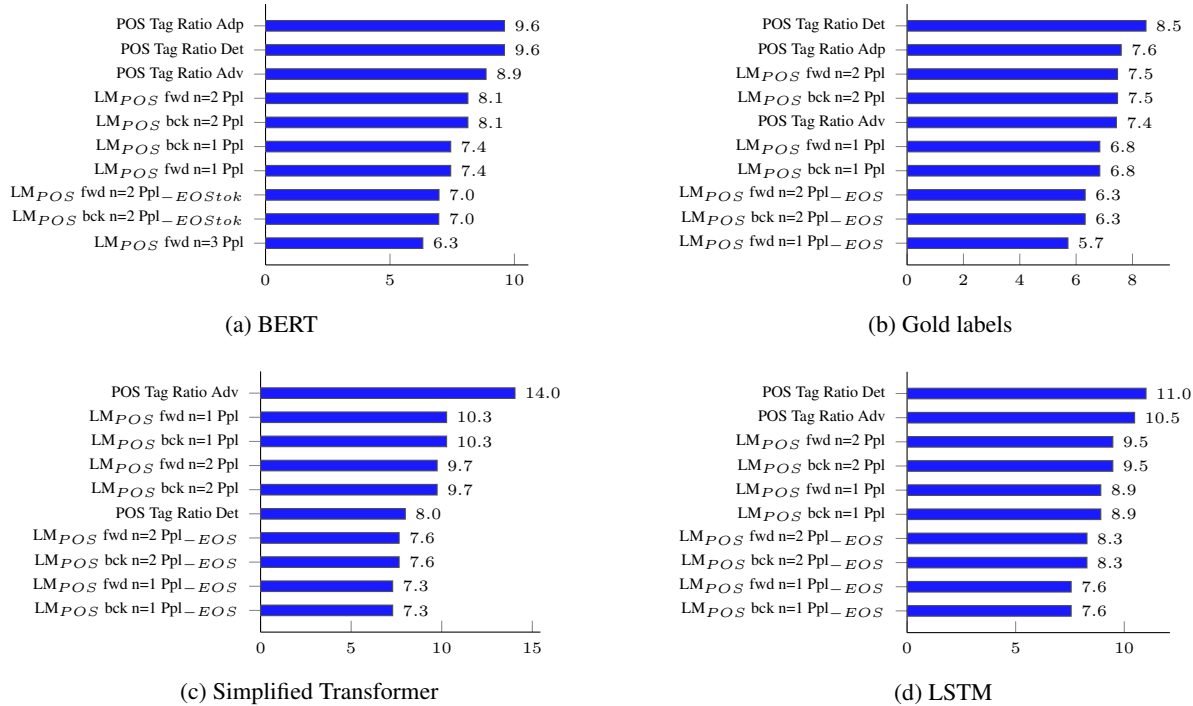

The $R^2$ measure is further used to rank features based on the amount of explained variance in predictions of a model. The top 10 features for predicting the predictions of each neural model and for predicting the actual gold labels are displayed in Figure~\ref{fig:plot_custom}. The order of top features is similar across the neural models (pairwise rank correlations $\rho_{Spearman}$ of at least 0.76), and similar to, but not identical to, the gold label results (pairwise rank correlations $\rho_{Spearman}$ of at least 0.75). We observe that most of the top features are either POS-perplexity-based, or bag-of-POS features. These features characterize interference in translationese. It also appears that more importance is attached to perplexities based on unigrams and bigrams than on other $n$-grams. Notably, the order of feature importance for the neural models is highly dissimilar from the order of hand-crafted feature weights for the SVM (pairwise rank correlations $\rho_{Spearman}$ at most 0.23). This might be connected to an accuracy gap between these models.

We conclude that many of hand-crafted translationese features are statistically significant for predicting the predictions of the neural models (and actual gold labels). However, due to the low $R^2$ values, we cannot conclude that the hand-crafted features explain the features learnt by the representation-learning models.

\section{Summary and Conclusion}
\label{conclusion}

This paper presents a systematic comparison of the performance of feature-engineering-based and feature-learning-based models on binary translationese classification tasks in various settings, i.e., monolingual single-source data, monolingual multi-source data, and multilingual multi-source data. Additionally, we analyze neural architectures to see how well the hand-crafted features explain the variance in the predictions of neural models. The results obtained in our experiments show that, $(i)$ representation-learning-based approaches outperform hand-crafted linguistically inspired feature-selection methods for translationese classification on a wide range of tasks, $(ii)$ the features learned by feature-learning based methods generalise better to different multilingual tasks and $(iii)$ our multilingual experiments provide empirical support for the existence of language independent translationese features. We also examine multiple neural architectures and confirm that translationese classification requires deep neural models for optimum results. We have shown that many traditional hand-crafted translationese features significantly predict the output of representation learning models, but may not necessarily explain their performance due to the weak correlation. Our experiments also show that even though single-source monolingual models yield the best performance, they can, to a reasonable extent, be substituted by multi-source mono- and multi-lingual models.

Our interpretability experiment provides only some initial insight into the neural models' performance. Even though there are significant relationships between many of the features and the neural models' predicted probabilities, further experiments are required to verify that the neural models actually use something akin to these features. Also our current approach ignores interaction between the features. In the future, we plan to conduct a more detailed analysis of the neural models' decision making.

\section*{Acknowledgements}

We would like to thank the reviewers for their insightful comments and feedback. This research is funded by the German Research Foundation (Deutsche Forschungsgemeinschaft) under
grant SFB 1102: Information Density and Linguistic Encoding.

\bibliography{anthology,custom}
\bibliographystyle{acl_natbib}

\clearpage

\appendix

\onecolumn
\section{Appendix}
\label{sec:appendix}

\subsection{List of hand-crafted features}
\label{subsec:feat_list}

\hspace{\parindent} 

\textbf{Col. 3:} SVM feature importance ranks (ranked by  absolute feature weight) for the model trained on the ALL--ALL[3] set.

\textbf{Col. 4-7:} Statistical significance of the features as predictors in per-feature linear regression with respect to neural models' predicted probabilities and gold labels (1 --significant with 99.9\% confidence level) on the ALL--ALL[3] test set.

\begin{center}
\begin{longtable}{|c|l|c|c|c|c|c|}

\hline \multicolumn{1}{|c|}{\textbf{ID}} & \multicolumn{1}{c|}{\textbf{Feature name}} & \multicolumn{1}{|p{1.5cm}|}{\centering{\textbf{SVM rank}}} &
\multicolumn{1}{|p{2.5cm}|}{\centering{\textbf{Simplified Transformer}}} &
\multicolumn{1}{c|}{\textbf{LSTM}} & \multicolumn{1}{c|}{\textbf{BERT}} & \multicolumn{1}{|p{1.5cm}|}{\centering{\textbf{Gold labels}}} \\ \hline 
\endfirsthead

\multicolumn{7}{c}%
{{\bfseries \tablename\ \thetable{} -- continued from previous page}} \\
\hline \multicolumn{1}{|c|}{\textbf{ID}} & \multicolumn{1}{c|}{\textbf{Feature name}} & \multicolumn{1}{|p{1.5cm}|}{\centering{\textbf{SVM rank}}} &
\multicolumn{1}{|p{2.5cm}|}{\centering{\textbf{Simplified Transformer}}} &
\multicolumn{1}{c|}{\textbf{LSTM}} & \multicolumn{1}{c|}{\textbf{BERT}} & \multicolumn{1}{|p{1.5cm}|}{\centering{\textbf{Gold labels}}}\\ \hline 
\endhead

\hline \multicolumn{7}{|r|}{{Continued on next page}} \\ \hline
\endfoot

\hline \hline
\endlastfoot
\hline

 \multicolumn{7}{|l|}{\textbf{Surface features}} \\ \hline
0 & Average word length & 11 & 1 & 1 & 1 & 1 \\ \hline
1 & Syllable ratio & 54 & 1 & 1 & 1 & 1 \\ \hline
2 & Paragraph length & 1 & 0 & 1 & 1 & 1 \\ \hline
\multicolumn{7}{|l|}{\textbf{Lexical features}} \\ \hline
3 & Lexical density & 19 & 0 & 1 & 1 & 1 \\ \hline
4 & Type-token ratio & 16 & 0 & 0 & 0 & 0 \\ \hline
\multicolumn{7}{|l|}{\textbf{Unigram bag-of-POS}} \\ \hline
5 & POS Tag Ratio Adj & 78 & 0 & 1 & 1 & 1 \\ \hline
6 & POS Tag Ratio Adp & 40 & 1 & 1 & 1 & 1 \\ \hline
7 & POS Tag Ratio Adv & 7 & 1 & 1 & 1 & 1 \\ \hline
8 & POS Tag Ratio Aux & 83 & 0 & 0 & 0 & 0 \\ \hline
9 & POS Tag Ratio Cconj & 80 & 0 & 0 & 0 & 0 \\ \hline
10 & POS Tag Ratio Det & 18 & 1 & 1 & 1 & 1 \\ \hline
11 & POS Tag Ratio Intj & 76 & 0 & 0 & 1 & 1 \\ \hline
12 & POS Tag Ratio Noun & 15 & 1 & 1 & 1 & 1 \\ \hline
13 & POS Tag Ratio Num & 46 & 0 & 0 & 0 & 0 \\ \hline
14 & POS Tag Ratio Part & 89 & 0 & 0 & 0 & 0 \\ \hline
15 & POS Tag Ratio Pron & 55 & 1 & 0 & 0 & 0 \\ \hline
16 & POS Tag Ratio Propn & 26 & 0 & 1 & 1 & 1 \\ \hline
17 & POS Tag Ratio Punct & 53 & 1 & 0 & 0 & 0 \\ \hline
18 & POS Tag Ratio Sconj & 41 & 0 & 0 & 1 & 0 \\ \hline
19 & POS Tag Ratio Space & 88 & 0 & 0 & 0 & 0 \\ \hline
20 & POS Tag Ratio Sym & 66 & 0 & 0 & 0 & 0 \\ \hline
21 & POS Tag Ratio Verb & 24 & 0 & 1 & 1 & 1 \\ \hline
22 & POS Tag Ratio X & 2 & 1 & 1 & 1 & 1 \\ \hline
\multicolumn{7}{|l|}{\textbf{SRILM language modelling features}} \\ \hline
23 & LM$_{tok}$ fwd n=1 Log Prob & 3 & 0 & 1 & 1 & 1 \\ \hline
24 & LM$_{tok}$ fwd n=1 Ppl & 47 & 1 & 1 & 1 & 1 \\ \hline
25 & LM$_{tok}$ fwd n=1 Ppl$_{-EOS}$ & 85 & 1 & 1 & 1 & 1 \\ \hline
26 & LM$_{tok}$ fwd n=2 Log Prob & 20 & 0 & 1 & 1 & 1 \\ \hline
27 & LM$_{tok}$ fwd n=2 Ppl & 45 & 0 & 0 & 0 & 0 \\ \hline
28 & LM$_{tok}$ fwd n=2 Ppl$_{-EOS}$ & 96 & 0 & 0 & 0 & 0 \\ \hline
29 & LM$_{tok}$ fwd n=3 Log Prob & 10 & 0 & 1 & 0 & 0 \\ \hline
30 & LM$_{tok}$ fwd n=3 Ppl & 28 & 0 & 0 & 0 & 0 \\ \hline
31 & LM$_{tok}$ fwd n=3 Ppl$_{-EOS}$ & 101 & 0 & 0 & 0 & 0 \\ \hline
32 & LM$_{tok}$ fwd n=4 Log Prob & 14 & 0 & 1 & 0 & 0 \\ \hline
33 & LM$_{tok}$ fwd n=4 Ppl & 27 & 0 & 0 & 0 & 0 \\ \hline
34 & LM$_{tok}$ fwd n=4 Ppl$_{-EOS}$ & 100 & 0 & 0 & 0 & 0 \\ \hline
35 & LM$_{tok}$ fwd n=5 Log Prob & 57 & 0 & 1 & 0 & 0 \\ \hline
36 & LM$_{tok}$ fwd n=5 Ppl & 29 & 0 & 0 & 0 & 0 \\ \hline
37 & LM$_{tok}$ fwd n=5 Ppl$_{-EOS}$ & 102 & 0 & 0 & 0 & 0 \\ \hline
38 & LM$_{tok}$ bck n=1 Log Prob & 4 & 0 & 1 & 1 & 1 \\ \hline
39 & LM$_{tok}$ bck n=1 Ppl & 48 & 1 & 1 & 1 & 1 \\ \hline
40 & LM$_{tok}$ bck n=1 Ppl$_{-EOS}$ & 86 & 1 & 1 & 1 & 1 \\ \hline
41 & LM$_{tok}$ bck n=2 Log Prob & 17 & 0 & 1 & 1 & 1 \\ \hline
42 & LM$_{tok}$ bck n=2 Ppl & 52 & 0 & 0 & 0 & 0 \\ \hline
43 & LM$_{tok}$ bck n=2 Ppl$_{-EOS}$ & 95 & 0 & 0 & 0 & 0 \\ \hline
44 & LM$_{tok}$ bck n=3 Log Prob & 25 & 0 & 1 & 0 & 0 \\ \hline
45 & LM$_{tok}$ bck n=3 Ppl & 32 & 0 & 0 & 0 & 0 \\ \hline
46 & LM$_{tok}$ bck n=3 Ppl$_{-EOS}$ & 98 & 0 & 0 & 0 & 0 \\ \hline
47 & LM$_{tok}$ bck n=4 Log Prob & 30 & 0 & 1 & 0 & 0 \\ \hline
48 & LM$_{tok}$ bck n=4 Ppl & 31 & 0 & 0 & 0 & 0 \\ \hline
49 & LM$_{tok}$ bck n=4 Ppl$_{-EOS}$ & 97 & 0 & 0 & 0 & 0 \\ \hline
50 & LM$_{tok}$ bck n=5 Log Prob & 61 & 0 & 1 & 0 & 0 \\ \hline
51 & LM$_{tok}$ bck n=5 Ppl & 37 & 0 & 0 & 0 & 0 \\ \hline
52 & LM$_{tok}$ bck n=5 Ppl$_{-EOS}$ & 99 & 0 & 0 & 0 & 0 \\ \hline
53 & LM$_{POS}$ fwd n=1 Log Prob & 8 & 0 & 1 & 1 & 1 \\ \hline
54 & LM$_{POS}$ fwd n=1 Ppl & 33 & 1 & 1 & 1 & 1 \\ \hline
55 & LM$_{POS}$ fwd n=1 Ppl$_{-EOS}$ & 58 & 1 & 1 & 1 & 1 \\ \hline
56 & LM$_{POS}$ fwd n=2 Log Prob & 5 & 0 & 1 & 0 & 0 \\ \hline
57 & LM$_{POS}$ fwd n=2 Ppl & 22 & 1 & 1 & 1 & 1 \\ \hline
58 & LM$_{POS}$ fwd n=2 Ppl$_{-EOS}$ & 93 & 1 & 1 & 1 & 1 \\ \hline
59 & LM$_{POS}$ fwd n=3 Log Prob & 12 & 0 & 1 & 1 & 1 \\ \hline
60 & LM$_{POS}$ fwd n=3 Ppl & 63 & 1 & 1 & 1 & 1 \\ \hline
61 & LM$_{POS}$ fwd n=3 Ppl$_{-EOS}$ & 106 & 1 & 1 & 1 & 1 \\ \hline
62 & LM$_{POS}$ fwd n=4 Log Prob & 43 & 0 & 1 & 1 & 1 \\ \hline
63 & LM$_{POS}$ fwd n=4 Ppl & 38 & 1 & 1 & 1 & 1 \\ \hline
64 & LM$_{POS}$ fwd n=4 Ppl$_{-EOS}$ & 104 & 1 & 1 & 1 & 1 \\ \hline
65 & LM$_{POS}$ fwd n=5 Log Prob & 75 & 0 & 1 & 1 & 1 \\ \hline
66 & LM$_{POS}$ fwd n=5 Ppl & 35 & 1 & 1 & 1 & 1 \\ \hline
67 & LM$_{POS}$ fwd n=5 Ppl$_{-EOS}$ & 103 & 1 & 1 & 1 & 1 \\ \hline
68 & LM$_{POS}$ bck n=1 Log Prob & 9 & 0 & 1 & 1 & 1 \\ \hline
69 & LM$_{POS}$ bck n=1 Ppl & 34 & 1 & 1 & 1 & 1 \\ \hline
70 & LM$_{POS}$ bck n=1 Ppl$_{-EOS}$ & 59 & 1 & 1 & 1 & 1 \\ \hline
71 & LM$_{POS}$ bck n=2 Log Prob & 6 & 0 & 1 & 0 & 0 \\ \hline
72 & LM$_{POS}$ bck n=2 Ppl & 23 & 1 & 1 & 1 & 1 \\ \hline
73 & LM$_{POS}$ bck n=2 Ppl$_{-EOS}$ & 94 & 1 & 1 & 1 & 1 \\ \hline
74 & LM$_{POS}$ bck n=3 Log Prob & 13 & 0 & 1 & 1 & 1 \\ \hline
75 & LM$_{POS}$ bck n=3 Ppl & 68 & 1 & 1 & 1 & 1 \\ \hline
76 & LM$_{POS}$ bck n=3 Ppl$_{-EOS}$ & 108 & 1 & 1 & 1 & 1 \\ \hline
77 & LM$_{POS}$ bck n=4 Log Prob & 42 & 0 & 1 & 1 & 1 \\ \hline
78 & LM$_{POS}$ bck n=4 Ppl & 51 & 1 & 1 & 1 & 1 \\ \hline
79 & LM$_{POS}$ bck n=4 Ppl$_{-EOS}$ & 107 & 1 & 1 & 1 & 1 \\ \hline
80 & LM$_{POS}$ bck n=5 Log Prob & 67 & 0 & 1 & 1 & 1 \\ \hline
81 & LM$_{POS}$ bck n=5 Ppl & 44 & 1 & 1 & 1 & 1 \\ \hline
82 & LM$_{POS}$ bck n=5 Ppl$_{-EOS}$ & 105 & 1 & 1 & 1 & 1 \\ \hline
\multicolumn{7}{|l|}{\textbf{N-gram freq. quartile distribution features}} \\ \hline
83 & \% unigrams from freq. quartile 1 & 50 & 0 & 0 & 1 & 1 \\ \hline
84 & \% unigrams from freq. quartile 2 & 39 & 1 & 1 & 1 & 1 \\ \hline
85 & \% unigrams from freq. quartile 3 & 65 & 0 & 0 & 0 & 0 \\ \hline
86 & \% unigrams from freq. quartile 4 & 90 & 1 & 1 & 1 & 1 \\ \hline
87 & \% OOV unigrams & 21 & 1 & 1 & 1 & 1 \\ \hline
88 & \% bigrams from freq. quartile 1 & 36 & 1 & 1 & 0 & 0 \\ \hline
89 & \% bigrams from freq. quartile 2 & 79 & 1 & 0 & 1 & 1 \\ \hline
90 & \% bigrams from freq. quartile 3 & 70 & 0 & 0 & 0 & 0 \\ \hline
91 & \% bigrams from freq. quartile 4 & 60 & 0 & 0 & 0 & 0 \\ \hline
92 & \% OOV bigrams & 69 & 0 & 0 & 0 & 0 \\ \hline
93 & \% trigrams from freq. quartile 1 & 77 & 0 & 0 & 0 & 0 \\ \hline
94 & \% trigrams from freq. quartile 2 & 49 & 0 & 0 & 0 & 0 \\ \hline
95 & \% trigrams from freq. quartile 3 & 56 & 0 & 0 & 0 & 0 \\ \hline
96 & \% trigrams from freq. quartile 4 & 64 & 0 & 0 & 0 & 0 \\ \hline
97 & \% OOV trigrams & 81 & 0 & 0 & 0 & 0 \\ \hline
98 & \% 4-grams from freq. quartile 1 & 82 & 0 & 0 & 1 & 1 \\ \hline
99 & \% 4-grams from freq. quartile 2 & 73 & 0 & 0 & 0 & 0 \\ \hline
100 & \% 4-grams from freq. quartile 3 & 72 & 0 & 0 & 0 & 0 \\ \hline
101 & \% 4-grams from freq. quartile 4 & 74 & 0 & 0 & 0 & 0 \\ \hline
102 & \% OOV 4-grams & 62 & 0 & 0 & 0 & 0 \\ \hline
103 & \% 5-grams from freq. quartile 1 & 84 & 0 & 0 & 0 & 0 \\ \hline
104 & \% 5-grams from freq. quartile 2 & 87 & 0 & 0 & 0 & 0 \\ \hline
105 & \% 5-grams from freq. quartile 3 & 91 & 0 & 0 & 0 & 0 \\ \hline
106 & \% 5-grams from freq. quartile 4 & 92 & 0 & 0 & 0 & 0 \\ \hline
107 & \% OOV 5-grams & 71 & 0 & 0 & 0 & 0
\end{longtable}
\end{center}

\clearpage
\twocolumn
\subsection{Simplified Transformer}
\label{subsec:sformer}

The simplified transformer differs from the standard transformer in the following ways:
\begin{enumerate}
    \item A cumulative sum-based contextualisation layer is used instead of positional encodings.
    \item The attention computation is reduced to element-wise operations and has no feedforward connections.
\end{enumerate}

\subsubsection{Encoder}
The encoder consists a contextualisation and attention layer with residual connections between both sublayers followed by layer normalisation.

\subsubsection{Contextualisation}
Given an input sequence $S = {s_1, s_2, ..., s_L}$, we obtain an embeddings matrix $X \in \mathbb{R}^{D \times L}$. The embeddings matrix $X$ is fed into the contextualisation layer of the transformer to obtain contextual embeddings $\hat{X} \in \mathbb{R}^{D \times L}$. We begin by taking a cumulative sum of the sequence of the embeddings $X$ as the context matrix
\begin{equation}
C = \sum_{j=1}^{L} X_{:j}
\end{equation}
followed by a column-wise dot product between the embeddings matrix ($X$) and the generated context ($C$) to get weights $w \in \mathbb{R}^{1 \times L}$:
\begin{equation}
w_{j} = X_{j} \cdot C_{j}
\end{equation}
where $j$ is the position of a word in the sequence $X$, $w$ is a row vector and $w_{j}$ is a scalar at position ${j}$. The original embeddings matrix ($X$) is then multiplied element-wise with the weights $w$ to obtain a contextualised representation of the sequence ($\hat{X}$):
\begin{equation}
    \hat{X} = X_{ij} \cdot w_{j}
\end{equation}

\subsubsection{Attention}
The attention takes 3 inputs: query, key and value. The output of the contextualisation layer ($\hat{X}$) is fed in as both the query and value. The context matrix ($C$) is fed in as the key. The query and key are passed through a feature map to obtain Q and K respectively. The feature map $(8)$ ensures Q and K are always positive. Therefore we can simplify softmax to the first term in the product in $(13)$: the Energy simply scaled by the sum of the Energies. The attention computation is formalised as follows:
\begin{equation}
    Feature\ map(x) = gelu(x) + 1
\end{equation}
\begin{equation}
    Q = |Feature\ map(query)|
\end{equation}
\begin{equation}
    K = |Feature\ map(key)|
\end{equation}
\begin{equation}
    V = value
\end{equation}
\begin{equation}
    Energy(E) = \frac{Q_{ij} \cdot K_{ij}}{\sqrt[]{D}}
\end{equation}
\begin{equation}
    Attention(A) = \frac{E_{j}}{\sum_{j=1}^{L}{E_{j}}} \cdot V
\end{equation}

\subsubsection{Decoder}
The decoder consists of two blocks. The first block is similar to the encoder block with a contextualisation layer and attention layer with residual connections between both sublayers followed by layer normalisation. The second block is another attention layer with residual connection to the previous block followed by a layer normalisation. In the second block, the output of the first block is fed in as the query and the output of the encoder block is fed in as the value. The key is the sum of the encoder's embedding matrix ($X$). This sum operation can be skipped by taking the last column of the encoder's context matrix ($C$). The decoder output ($Y \in \mathbb{R}^{D \times L}$) is pooled (average) and the resulting $D \times 1$ vector is passed to a classifier.

\clearpage
\subsection{Handcrafted Feature-based SVM Baseline}
\label{subsec:baselines}
For the POS-trigrams and character-trigrams baselines we implement the setup from \citet{rabinovichWintner:2015} (and, respectively, \citet{volanskyEtAl:2015}). In both cases we only take 1000 most frequent trigrams. The values correspond to the relative frequency of the trigram in the paragraph normalized by the total number of trigrams in the paragraph. For POS-trigrams, we pad the paragraphs with special start-of-string and end-of-string tokens. For character trigrams, we pad each word in this way, and avoid cross-token trigrams, as well as punctuation. Results are displayed in Table \ref{Baselines}.

Rabinovich and Wintner (2015) report much higher accuracy (> 90) on their Europarl data: they classify text chunks of 2000 tokens, while we report results on paragraphs with average length of around 80 tokens.

\begin{table}[!htbp]
\small
\centering
\begin{tabular}{lcc}
\toprule
& \texttt{POS-trigrams} & \texttt{char-trigrams}   \\ \midrule
DE--EN & 76.6$\pm$0.0 ($\uparrow$5.1) & 67.5$\pm$0.0 ($\downarrow$4.1) \\
DE--ES & 80.3$\pm$0.0 ($\uparrow$4.1) & 71.2$\pm$0.0 ($\downarrow$5.0) \\
EN--DE & 73.1$\pm$0.0 ($\uparrow$5.5) & 62.9$\pm$0.0 ($\downarrow$4.8) \\
EN--ES & 73.4$\pm$0.0 ($\uparrow$3.4) & 66.1$\pm$0.0 ($\downarrow$4.0) \\
ES--DE & 76.1$\pm$0.0 ($\uparrow$5.1) & 66.6$\pm$0.0 ($\downarrow$4.4) \\
ES--EN & 74.2$\pm$0.0 ($\uparrow$7.6) & 64.4$\pm$0.0 ($\downarrow$2.3) \\
DE--ALL & 76.7$\pm$0.0 ($\uparrow$4.1) & 68.4$\pm$0.0 ($\downarrow$4.2) \\
EN--ALL & 69.7$\pm$0.0 ($\uparrow$4.4) & 62.1$\pm$0.0 ($\downarrow$3.3) \\
ES--ALL & 72.9$\pm$0.0 ($\uparrow$5.5) & 63.3$\pm$0.0 ($\downarrow$4.1) \\
ALL--ALL[3] & 66.6$\pm$0.0 ($\uparrow$7.6) & 62.2$\pm$0.1 ($\uparrow$3.3) \\
ALL--ALL[8] & 61.6$\pm$0.0 ($\downarrow$3.8) & 61.5$\pm$0.0 ($\downarrow$3.9) \\
\bottomrule
\end{tabular}
\caption{\label{Baselines} Test accuracy of baseline systems implemented from \cite{rabinovichWintner:2015}. The mean and the standard deviations over 5 runs are reported. The difference from the \texttt{Handcr.+SVM} model is indicated in parentheses.
}
\end{table}

\end{document}